\documentclass{article}
\usepackage{spconf,amsmath,graphicx}
\usepackage{amssymb, amsmath}
\usepackage{url}
\usepackage{caption, subcaption}
\usepackage{float}

\usepackage{tikz}
\usetikzlibrary{shapes.geometric, arrows}
\tikzstyle{arrow} = [thick,->,>=stealth]
\usetikzlibrary{shapes.multipart}
\tikzset{state/.style={rectangle split, rectangle split parts=2, rectangle split part fill={pink!30,teal!20}, rounded corners, draw=black, minimum height=1.5cm, text width=3cm, inner sep=2pt, text centered,}}
\tikzstyle{io} = [rectangle, rounded corners, minimum width=1.25cm, minimum height=0.75cm, text centered, draw=black, fill=blue!30]
\tikzstyle{blank} = [rectangle, rounded corners, minimum width=1.25cm, minimum height=0.75cm, text centered, draw=white, fill=white!30]
\tikzstyle{cnn} = [rectangle, rounded corners, minimum width=1.25cm, minimum height=0.75cm, text centered, draw=black, fill=green!30]
\tikzstyle{rnn} = [rectangle, rounded corners, minimum width=1.25cm, minimum height=0.75cm, text centered, draw=black, fill=yellow!30]
\tikzstyle{real} = [rectangle, rounded corners, minimum width=1.25cm, minimum height=0.75cm, text centered, draw=black, fill=pink!30]
\tikzstyle{imag} = [rectangle, rounded corners, minimum width=1.25cm, minimum height=0.75cm, text centered, draw=black, fill=teal!30]

\title{Denoising neural networks for magnetic resonance spectroscopy}

\name{Natalie Klein$^{\star}$ \quad Amber J. Day$^{\star \dagger}$ \quad Harris Mason$^{\star}$ \quad Michael W. Malone$^{\star}$ \quad Sinead A. Williamson$^{\dagger}$
}

\address{$^{\star}$Los Alamos National Laboratory, Los Alamos, NM 87545, USA\\
         $^{\dagger}$Department of Statistics and Data Sciences, University of Texas at Austin, Austin, TX 78712, USA}

\begin{document}
\topmargin=0mm
%
\maketitle

\begin{abstract}
In many scientific applications, measured time series are corrupted by noise or distortions.
Traditional denoising techniques often fail to recover the signal of interest, particularly when the signal-to-noise ratio is low or when certain assumptions on the signal and noise are violated.
In this work, we demonstrate that deep learning-based denoising methods can outperform traditional techniques while exhibiting greater robustness to variation in noise and signal characteristics.
Our motivating example is magnetic resonance spectroscopy, in which a primary goal is to detect the presence of short-duration, low-amplitude radio frequency signals that are often obscured by strong interference that can be difficult to separate from the signal using traditional methods. 
We explore various deep learning architecture choices to capture the inherently complex-valued nature of magnetic resonance signals.
On both synthetic and experimental data, we show that our deep learning-based approaches can exceed performance of traditional techniques, providing a powerful new class of methods for analysis of scientific time series data.
\end{abstract}
\begin{keywords}
Deep learning, signal denoising, complex-valued neural networks, nuclear quadrupole resonance
\end{keywords}
%

\section{Introduction}
\label{sec:intro}
In this work, we consider deep learning-based time-domain denoising of single-sensor complex-valued (CV) signals with expected variation in signal and noise properties during deployment.
Our motivating application is magnetic resonance spectroscopy, a chemical analysis technique that can be used to identify materials and characterize the structure and chemical environment of molecules in the sample~\cite{miller2007nuclear}. 
We focus specifically on nuclear quadrupole resonance (NQR) spectroscopy, which can capture unique signatures arising from transitions of nuclei in the absence of an applied magnetic field and can thus provide a sensitive standoff detection capability for some classes of substances~\cite{garroway2001remote}. 
Our ultimate goal is to extract clean signals from noisy NQR recordings of a given substance for downstream tasks such as signal classification.

\begin{figure}[ht]
    \centering
    \includegraphics[width=0.9\linewidth]{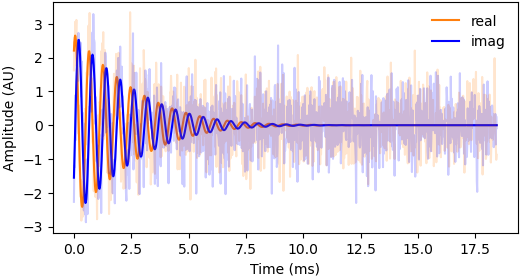}
    \caption{Simulated clean signal (dark lines), with real (orange) and imaginary (blue) components, overlaid on noisy signal (light lines) generated by combining actual recordings of radio frequency interference with the clean signal. Noise and distortions of the signal based on environmental conditions often complicate signal detection and characterization.}
    \label{fig:example_data}
\end{figure}

In practice, many substances of interest produce very low-amplitude, short-duration radio frequency (RF) signals that are difficult to detect due to noise and environmental interference, including commonly-occurring RF interference (illustration: Fig.~\ref{fig:example_data}). 
In addition, expected signal characteristics may vary during deployment; for example, temperature changes can shift the expected signal frequency~\cite{koukoulas1990observations} and crystalline samples with strong dipolar coupling can deviate from theoretical exponential decay models~\cite{ramani1976fourier}.
Existing signal processing methods (e.g., wavelet decomposition, time-domain deconvolution, or subspace methods) have been found to work well when there is a high signal-to-noise ratio (SNR), but we show that their performance degrades in more challenging situations (e.g., low SNR, non-white noise, or overlapping signal and noise frequency components). 

Neural networks (NNs) have been used to isolate signals of interest across a variety of applications, including image denoising~\cite{vincent2010stacked} and speech separation~\cite{luo2019conv}.
We adapt such approaches to the task of NQR denoising, incorporating CV weights and activation functions to avoid losing information in the CV signals.
We show that NN-based approaches outperform existing denoising methods, particularly in low SNR scenarios, with CV-NNs providing more robust results across changes in noise structure and signal power.

\section{RELATION TO PRIOR WORK}
\label{sec:prior}
NQR denoising methods seek to recover clean signals from noisy signals for improved chemical detection and characterization; we consider a single-sensor setup in which it can be particularly difficult to separate noise and RF interference from the signal.
In addition, the short-duration signals often exhibit low SNR, and signals cannot be expected to conform to simple physical signal models (such as complex exponential decay with a known frequency).
Previous approaches for signal denoising in NQR include both time-domain and frequency-domain denoising methods and interference cancellation (see~\cite{monea2021signal} for a review), with some methods leveraging specialized collection techniques with multiple sensors. 
For the single-sensor scenario, we investigate automated versions of three existing methods: wavelet denoising (e.g.,~\cite{shanemd}), statistical curve fitting of assumed signal models (e.g.,~\cite{somasundaram2008countering}), and subspace separation methods (e.g.,~\cite{varma2021novel}).
Recently, real-valued (RV) denoising autoencoders (AEs) were shown to improve signal detection when applied to low SNR NQR data for denoising prior to classification~\cite{monea2022enhancing}.

Our work differs from~\cite{monea2022enhancing} in two key ways.
First, we make use of CV-NNs, allowing native processing of CV signals and improving learning by exploiting structured relationships between real and imaginary components~\cite{trabelsi1705deep}.
Several recent works suggest CV-NNs can outperform RV-NNs (e.g.,~\cite{watcharasupat2022end}), though this may be context-dependent~\cite{hirose2012generalization}.
Second, we compare denoising AEs to a more sophisticated and specialized architecture designed for speech separation, ConvTasNet (CTN)~\cite{luo2019conv}.
This fully-convolutional network utilizes an encoder and decoder, similar to an AE, but explicitly learns to separate multiple sources (here, signal and noise) with a masking module.

\section{Proposed denoising neural networks}
\label{sec:proposed}
We seek to compare RV and CV variants of two NN approaches, AE and CTN, against existing denoising methods, and to understand which NNs provide more accurate and robust results in this application. Full implementation details are available in our GitHub repository~\cite{Klein_SplitML_2022}.

\subsection{Neural network (NN) architectures}
Our CTN implementation~\cite{Klein_SplitML_2022} is based on the \texttt{torchaudio} implementation~\cite{yang2022torchaudio}, but modified to include three-layer deep encoder/decoder modules~\cite{kishore2020improved} and to accommodate CV data (Section \ref{subsec:complex_nn}).
We fixed most CTN hyperparameters to default values in the \texttt{torchaudio} implementation, but select among three window lengths (32, 64, or 128 samples) based on test set performance.
For the AE model, we use a fully-connected network with encoder consisting of one intermediate hidden layer (dimension 10) that maps to a five-dimensional latent representation.
Models are trained using signal reconstruction error.
For each considered model, we train an ensemble of models in which the weights are randomly initialized with different random seeds~\cite{lakshminarayanan2017simple}; this provides a metric of uncertainty and the opportunity to average across model predictions for higher accuracy.

\subsection{Complex-valued (CV) and real-valued (RV) NNs}
\label{subsec:complex_nn}
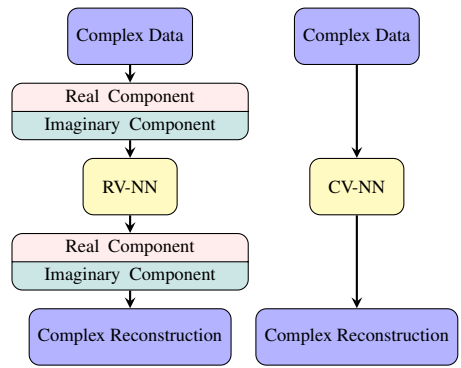
\begin{figure}[ht]
\centering
    \begin{tikzpicture}[node distance=1cm]
    <TikZ code>
    \node (in1) [io] {\scriptsize Complex Data};
    \node (conc1) [state, below of=in1] {\scriptsize Real Component \nodepart{two} \scriptsize Imaginary Component};
    \node (nn1) [rnn, below of=conc1] {\scriptsize RV-NN};
    \node (conc2) [state, below of=nn1] {\scriptsize Real Component \nodepart{two} \scriptsize Imaginary Component};
    \node (out1) [io, below of=conc2] {\scriptsize Complex Reconstruction};
    \draw [arrow] (in1) -- (conc1);
    \draw [arrow] (conc1) -- (nn1);
    \draw [arrow] (nn1) -- (conc2);
    \draw [arrow] (conc2) -- (out1);
    \end{tikzpicture}
    \begin{tikzpicture}[node distance=1cm]
    <TikZ code>
    \node (in1) [io] {\scriptsize Complex Data};
    \node (nn1) [rnn, below of=in1, node distance=2cm] {\scriptsize CV-NN};
    \node (out1) [io, below of=nn1, node distance=2cm] {\scriptsize Complex Reconstruction};
    \draw [arrow] (in1) -- (nn1);
    \draw [arrow] (nn1) -- (out1);
    \end{tikzpicture}
    \caption{The first approach to NNs for CV data (left) jointly processes real and imaginary components of the data with a real-valued neural network (RV-NN), while the second approach (right) operates directly on complex data with a complex-valued neural network (CV-NN).} \label{fig:arch}
\end{figure}

We explore two main approaches to training NNs for CV data: one that relies on an RV-NN to operate on separated real and imaginary components with learnable RV weights representing relationships between the components (Fig. \ref{fig:arch}, left side) and one that uses a CV-NN to directly operate on CV inputs (Fig. \ref{fig:arch}, right side). 
In~\cite{Klein_SplitML_2022}, we explore alternative RV approaches, but find they perform worse than the presented RV approach.
For the CV-NN approach, we mirror the RV-NN architecture, but with CV activation functions and convolutions suitable for complex data. 

\section{Experimental Results}
To evaluate performance, we generate a set of synthetic NQR datasets with variation in SNR, signal model, and noise source (random white or actual measured RF interference). 
We compare the NNs to each other and to three baseline methods via a scaled test set $R^2$ metric, then evaluate robustness of the NNs to out-of-distribution data.
Finally, we apply the best-performing methods to experimental data.

\subsection{Datasets}
For NN training and quantitative evaluation, we utilize synthetic signals generated according to a physics-based model.
Specifically, time-varying signals $y(t)$ are generated as a noisy decaying complex exponential (Voigt profile~\cite{ramani1976fourier}),
\begin{align}
    y(t) = A \exp\left(- \frac{t^2}{2\sigma^2}-\frac{t}{T_2} + i[2\pi w t + \phi]\right) + \epsilon(t), \label{eq:datagen}
\end{align}
where $i$ is the imaginary unit, $A$ is the initial amplitude, $\sigma$ and $T_2$ control the rate of decay, $\phi$ is the phase offset, $w$ is the frequency, and $\epsilon$ is additive noise.
We draw $\sigma$ and  $T_2$ from $\text{Unif}(0.001, 0.01)$ to emulate decay rate variation relative to recording length that might be observed in practice.
The parameter $\sigma \ll \infty$ models expected deviations from simple exponential decay.
We generate eight training datasets by selecting from two noise distributions, two signal frequency bands centered at $f_0=0$Hz and $f_0=F=1600$Hz, and two SNR regimes (low and high).
Varying the frequency as $\text{Unif}(f_0-150, f_0+150)$ emulates temperature-dependent signal frequency variation that might be encountered in practice.
The SNRs range between $[-36\text{dB}, -7\text{dB}]$  and $[-10\text{dB}, 3\text{dB}]$ for low and high SNR cases, respectively, with average low and high SNRs at approximately -17dB and -3dB.
The noise distributions are random Gaussian white noise and measured RF noise from a laboratory instrument with no signal present; we note that $F=1600$Hz was chosen because the measured noise features elevated power at this frequency, leading to a more challenging noise separation task.
For each dataset, we train an independent ensemble of NNs using 6,000 noisy signals for training and two independent sets of 2,000 noisy signals for validation and testing.
We also apply our methods to representative experimental data exhibiting relatively high SNR and signal frequency near 0Hz (due to complex demodulation at a known signal frequency).
For more complete descriptions of each dataset, see~\cite{Klein_SplitML_2022}.

\subsection{Comparison of NN architectures}
First, we compare test set ensemble performance across the eight datasets, two model architectures (AE and CTN), and two approaches to CV data (CV-NN and RV-NN) using a scaled $R^2$ metric, $R^2 = 100 \times \left( 1 - \frac{\sum_i \sum_t (y_i(t) - \hat{y}_i(t))^2}{\sum_i \sum_t y_i(t)^2 } \right)$, where $y_i(t)$ is the true clean signal instance $i$ at time $t$ and $\hat{y}_i(t)$ is the corresponding denoised signal prediction. 
The best possible performance would occur at $R^2 = 100.0$, with lower (or even negative) values indicating poor performance; the metric can be interpreted as measuring improvement of a prediction above a simple prediction of ``no signal present''.
Table \ref{tab:nn_r2} shows the results for the best performing AE and CTN models; overall, CV-CTN outperforms both RV-CTN and the AEs, though CV-CTN and RV-CTN achieve similar performance in high SNR.
Interestingly, in low SNR, CV-CTN performs better on measured noise than white noise, perhaps because it learns some relevant structure in the noise.

\begin{table}[ht]
    \centering
    \setlength\tabcolsep{2.5pt}
    \begin{subtable}[t]{\columnwidth}
        \caption{High SNR regime NN scaled $R^2$ values.}\label{subtab-1:nn_r2}
        \centering
        \begin{tabular}{r|r|r|r|r||r|r|r|r|}
           \multicolumn{1}{c}{} & \multicolumn{4}{c}{White noise} & \multicolumn{4}{c}{Measured noise} \\ \cline{2-9}
           & \multicolumn{2}{c|}{$f \sim 0$} & \multicolumn{2}{c||}{$f \sim F$} & \multicolumn{2}{c|}{$f \sim 0$} & \multicolumn{2}{c|}{$f \sim F$} \\ \hline
           Model & \multicolumn{1}{c|}{CV} & \multicolumn{1}{c|}{RV} & \multicolumn{1}{c|}{CV} & \multicolumn{1}{c||}{RV} & \multicolumn{1}{c|}{CV} & \multicolumn{1}{c|}{RV} & \multicolumn{1}{c|}{CV} & \multicolumn{1}{c|}{RV} \\ \hline
           AE  &  96.0 &  98.1     & 96.1       &95.5& 95.5& 98.1& 95.2& 95.8 \\
           CTN & {\bf 99.4} & \textbf{99.4} & {\bf 99.5} &\textbf{99.5} & 99.5 & {\bf 99.6} & {\bf 99.5} & \textbf{99.5} \\
            \hline
        \end{tabular}
    \end{subtable}
    \begin{subtable}[t]{\columnwidth}
        \caption{Low SNR regime NN scaled $R^2$ values.}\label{subtab-2:nn_r2}
        \centering
        \begin{tabular}{r|r|r|r|r||r|r|r|r|}
           \multicolumn{1}{c}{} & \multicolumn{4}{c}{White noise} & \multicolumn{4}{c}{Measured noise} \\ \cline{2-9}
           & \multicolumn{2}{c|}{$f \sim 0$} & \multicolumn{2}{c||}{$f \sim F$} & \multicolumn{2}{c|}{$f \sim 0$} & \multicolumn{2}{c|}{$f \sim F$} \\ \hline
           Model & \multicolumn{1}{c|}{CV} & \multicolumn{1}{c|}{RV} & \multicolumn{1}{c|}{CV} & \multicolumn{1}{c||}{RV} & \multicolumn{1}{c|}{CV} & \multicolumn{1}{c|}{RV} & \multicolumn{1}{c|}{CV} & \multicolumn{1}{c|}{RV} \\ \hline
           AE & 67.1 & 76.9 & 65.9 & 75.8 & 72.2 & 77.6 & 65.5 & 70.0  \\
           CTN & \textbf{81.7} & 81.1 &\textbf{ 81.8} & 81.2 & \textbf{85.5} & 84.9 &\textbf{ 83.7 }& 82.9 \\
            \hline
        \end{tabular}
    \end{subtable}
    \caption{Scaled $R^2$ (best possible value: 100.0) of NN ensemble test set predictions under different training and model scenarios. Within each SNR, AE and CTN results are split across noise distributions (white or measured), central signal frequency $f$, and complex-valued (CV) or real-valued (RV) NNs, with best results in each case bolded. Overall, CV-CTN performs the best within each noise, frequency, and SNR setting, particularly in low SNR. 
    }\label{tab:nn_r2}
\end{table}

\subsection{Comparison to baseline methods}
Next, we compare CV-CTN to three baseline methods: time-domain deconvolution (DC; in which we fit the signal model of Eq.~\ref{eq:datagen} to noisy data via nonlinear least squares~\cite{somasundaram2008countering,varma2021novel}), singular spectrum analysis (SSA; a subspace-based method~\cite{broomhead1986extracting,wang2021fast,varma2021novel}), and wavelet decomposition (W;~\cite{shanemd}).
Similar to NN-based approaches, each method uses automated optimization or thresholding rather than human intervention.
As shown in Table \ref{subtab-1:nn_v_baseline}, in the high SNR regime, many of the baseline methods perform nearly as well as CV-CTN, although SSA degrades under measured noise and wavelets degrade when $f \sim F$.
We note that initial values for DC are based on the known training distribution; performance of DC in practice is likely to be worse if poor initial values are used.
In low SNR (Table \ref{subtab-2:nn_v_baseline}), all of the baseline methods are considerably worse than CV-CTN.
These results suggest that compared to baseline methods, CV-CTN achieves good performance in low SNR conditions, with little dependence on the generating noise or signal properties.

\begin{table}[ht]
    \setlength\tabcolsep{2.5pt}
    \begin{subtable}[t]{\columnwidth}
    \caption{High SNR regime scaled $R^2$ values.}\label{subtab-1:nn_v_baseline}
        \centering
        \begin{tabular}{r|r|r||r|r|}
           \multicolumn{1}{c}{} & \multicolumn{2}{c}{White noise} & \multicolumn{2}{c}{Measured noise} \\ \cline{2-5}
           Method & \multicolumn{1}{c|}{$f \sim 0$} & \multicolumn{1}{c||}{$f \sim F$} & \multicolumn{1}{c|}{$f \sim 0$} & \multicolumn{1}{c|}{$f \sim F$} \\ \hline
           CV-CTN & \textbf{99.4} &\textbf{99.5} &\textbf{99.5}& \textbf{99.5} \\
           DC & 99.2 &99.1 &99.0&  \textbf{99.5} \\
           SSA & 98.6 & 98.6 & 52.4 & 52.6   \\
           W & 97.8& 87.0&  98.4& 89.0  \\
           \hline
        \end{tabular}
    \end{subtable}
    \begin{subtable}[t]{\columnwidth}
        \caption{Low SNR regime scaled $R^2$ values.}\label{subtab-2:nn_v_baseline}
        \centering
        \begin{tabular}{r|r|r||r|r|}
           \multicolumn{1}{c}{} & \multicolumn{2}{c}{White noise} & \multicolumn{2}{c}{Measured noise} \\ \cline{2-5}
           Method & \multicolumn{1}{c|}{$f \sim 0$} & \multicolumn{1}{c||}{$f \sim F$} & \multicolumn{1}{c|}{$f \sim 0$} & \multicolumn{1}{c|}{$f \sim F$} \\ \hline
           CV-CTN & \textbf{81.7} & \textbf{81.8} & \textbf{85.5} & \textbf{83.7} \\
           DC &  57.7& 65.5& 68.0&  62.4\\
           SSA & -42.6 &   -41.7 & -1970.2 & -1970.2  \\
           W & 21.5& -38.1&  51.5& -10.9 \\
           \hline
        \end{tabular}
    \end{subtable}
    \caption{Scaled $R^2$ of CV-CTN compared to time-domain deconvolution (DC), singular spectrum analysis (SSA), and wavelets (W). CV-CTN generally outperforms the other methods, and is consistent across signal and noise distributions. In contrast, the baseline methods are competitive with CV-CTN only in high SNR, and performance can depend on signal/noise characteristics (e.g., for SSA and W).}\label{tab:nn_v_baseline}
\end{table}

\subsection{Out-of-distribution performance}
Out-of-distribution (OOD) refers to the situation in which a machine learning model is applied to data that differs in distribution from the training data; in our application, OOD data could occur due to changes in noise/interference or changes in the signal during deployment.
While deep learning models generally do not perform as well on OOD data, ideally, useful NN denoising models would be somewhat robust to OOD data, or would at least provide uncertainty metrics to alert users when applied to OOD data.
We use two OOD tests: one based on changing the test set noise distribution relative to the training noise distribution, and one based on changing the test set signal frequency band relative to the training signal frequency band.
We omit full description of results due to space constraints (see~\cite{Klein_SplitML_2022}).
Briefly, we find that applying to test data with a different noise distribution does not appreciably degrade performance or increase the ensemble uncertainty.
However, significant changes to the signal distribution (e.g., training on $f = 0$Hz and applying to $f = 300$Hz) does increase error.
Here, the CV-CTN ensemble uncertainty increases under high SNR, suggesting measures of ensemble uncertainty could be used to identify OOD data, but the same pattern does not hold in low SNR. 
We conclude that for a given application, a suitable training set should cover the space of signal variations expected in practice.

\subsection{Application to experimental data}
\begin{figure}[ht]
    \begin{subfigure}[h]{\linewidth}
        \caption{Experimental data high SNR denoising results.}
        \label{fig:realdata_time}
        \centering
        \includegraphics[width=0.9\linewidth]{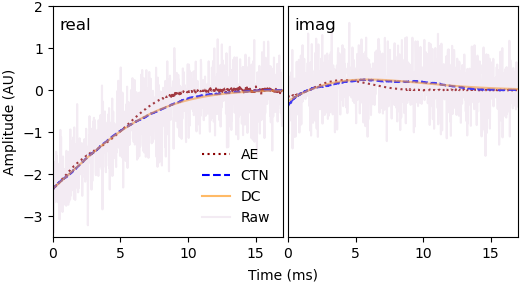}
        
    \end{subfigure}
    \begin{subfigure}[h]{\linewidth}
        \centering
        \caption{Synthetic data low SNR denoising results.}
        \label{fig:realdata_lowsnr}
        \includegraphics[width=0.9\linewidth]{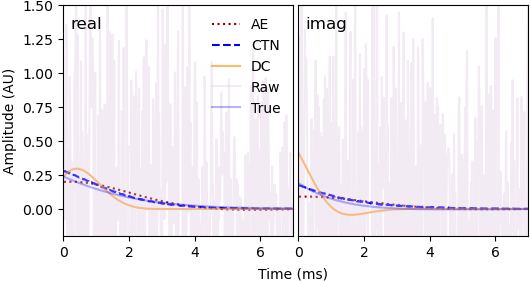}
    \end{subfigure}
    \caption{Three methods (RV-AE, CV-CTN, and DC) applied to (a) representative experimental data and (b) realistic, but low SNR, synthetic data. In (a), CV-CTN and DC agree closely, while RV-AE slightly deviates. In (b), CV-CTN recovers the ground truth signal shape more accurately than RV-AE or DC.}
    \label{fig:realdata}
\end{figure}

We apply the best-performing NNs (CV-CTN and RV-AE) to experimental data that exhibits signal frequency near zero and high SNR, suggesting networks trained on $f \sim 0$, high SNR data are suitable. 
Fig. \ref{fig:realdata_time} compares the NN predictions to the best high SNR baseline method (DC).
While no ground truth is available for this data, we note that DC and CV-CTN give very similar signal predictions, while RV-AE gives generally noisier predictions with distorted signal shape.
To better illustrate differences between the methods, we also show results for synthetic low SNR data with $f \sim 0$ (Fig. \ref{fig:realdata_lowsnr}).
Here, both RV-AE and DC deviate significantly from the true signal, while CV-CTN recovers it.
These results suggest that CV-CTN accurately recovers the signal in both SNRs.

\section{CONCLUSIONS}
In this work, we explored real- and complex-valued versions of two NN architectures for denoising of CV data. 
We found that among NNs, CV-CTN achieved the best denoising results.
However, some RV-NNs capable of learning relationships between real and imaginary components reached similar performance with reduced computational complexity.
We also demonstrated that NN approaches outperformed baseline denoising methods, particularly in low SNR. 
Our results also suggest that NN approaches are robust to some signal and noise variation, important for our intended application to magnetic resonance spectroscopy, where measurements are often corrupted by noise, interference, and distortions due to environmental conditions.
In future work, we intend to explore other NN architectures, including recurrent NNs, and the influence of the training set on the robustness of the resulting denoising network.
In addition, we plan to investigate the impact of our denoising approaches on downstream tasks, such as classification. 
Additional results and details are available on GitHub~\cite{Klein_SplitML_2022}.

\section{Acknowledgements}
This project was supported by the Laboratory Directed Research and Development program of Los Alamos National Laboratory under project number LDRD-20220086DR.
This manuscript has been approved for unlimited release and has been assigned LA-UR-22-30980. This work has been authored by an employee of Triad National Security, LLC which operates Los Alamos National Laboratory under Contract No. 89233218CNA000001 with the U.S. Department of Energy/National Nuclear Security Administration. The publisher, by accepting the article for publication, acknowledges that the United States Government retains a non-exclusive, paid-up, irrevocable, world-wide license to publish or reproduce the published form of the manuscript, or allow others to do so, for United States Government purposes.

\bibliographystyle{IEEEbib}
\bibliography{refs}

\clearpage
\onecolumn
\appendix

\renewcommand{\thetable}{\Alph{section}\arabic{table}}
\renewcommand{\thefigure}{\Alph{section}\arabic{figure}}
\section{Appendix for ``Denoising neural networks for magnetic resonance spectroscopy''}

\noindent Authors: \quad Natalie Klein$^{\star}$ \quad Amber J. Day$^{\star \dagger}$ \quad Harris Mason$^{\star}$ \quad Michael W. Malone$^{\star}$ \quad Sinead A. Williamson$^{\dagger}$\\
$^{\star}$Los Alamos National Laboratory, Los Alamos, NM 87545, USA\\
$^{\dagger}$Department of Statistics and Data Sciences, University of Texas at Austin, Austin, TX 78712, USA

\subsection{Dataset details}
\label{appendix:data}
When generating synthetic data based on Equation \ref{eq:datagen}, we use a time vector with 1,024 points at a sample spacing of $1.8 \times 10^{-5}$ seconds to correspond to the experimental data.
We draw parameters $T_2$ and $\sigma$ independently according to a Uniform($1 \times 10^{-3}, 1 \times 10^{-2})$ distribution.
Phase shift $\phi$ is drawn according to a Uniform($-\pi, \pi$) distribution.
The noise is drawn from one of two distributions: white Gaussian noise (`white') and experimental noise measurements from a nuclear quadrupole resonance (NQR) instrument with no relevant signal present (`measured'). 
The measured noise consists of 4,096 independent measurements of noise at the same sampling rate with duration 16,384 samples; we first split the independent measurements into training, validation, and test sets before dividing into pieces of length 1,024 to ensure no single measurement vector was reused in the validation or test sets.
We consider signals centered at frequency  $f \sim 0$Hz (to emulate complex-demodulated data) and signals centered at a higher frequency $f \sim F$, with $F = 1600$Hz (chosen to have signal power in a frequency band for which the measured noise has elevated power).
Given a choice of $f$, the frequency of an individual signal in the data set is drawn as Uniform($f-150, f+150$).
We control the signal-to-noise ratio (SNR) of the signals by varying $A$ relative to standardized noise (variance equal to one), with a low-SNR regime, $A \sim \text{Uniform}(0.1, 1.0)$, and a high-SNR regime, $A \sim \text{Uniform}(2.1, 3.0)$.
The experimental data comes from nuclear magnetic resonance measurements of NaNO$_2$.
Complex demodulation at the expected resonance frequency was performed during data collection, resulting in complex-valued time series data expected to be centered at $f\sim 0$Hz. It consists of the same time vector as the synthetically generated signal and upon inspection was seen to have high SNR, with visible exponential decay trend in the noisy signals.

\subsection{SNR across datasets}
Signal-to-noise ratio (SNR) is defined as the ratio of the power of a signal ($P_{signal}$) to the power of background noise ($P_{noise}$),  
    $\text{SNR} = \frac{P_{signal}}{P_{noise}}$.
We reference the logarithmic decibel scale of the SNR (SNR$_{dB}$) for ease of comparison:
    $\text{SNR}_{dB}=10 \log_{10}\left( \frac{P_{signal}}{P_{noise}}\right) =10 \log_{10}(\text{SNR})$.
Since the SNR$_{dB}$ values are approximately the same across noise distributions and between the in-distribution and out-of-distribution datasets, we have only reported the values for the white-noise and in-distribution datasets.

\begin{table}[H]
    \centering
\begin{tabular}{|l|l|c|c|c|c} \hline
 SNR Type & Frequency &        Minimum SNR$_{dB}$ &       Average SNR$_{dB}$ &       Maximum SNR$_{dB}$ \\ \hline
low & $f \sim 0$ & -35.6 & -17.7 & -7.3 \\ \hline
low & $f \sim 1600$ & -35.2 & -17.0 & -7.2 \\ \hline
high & $f \sim 0$ &  -10.0 &  -3.1 &  2.5 \\ \hline
high & $f \sim 1600$ &  -9.7 &  -2.4 &  2.6 \\ \hline
\end{tabular}
    \caption{Observed minimum, average, and maximum SNR (dB) in the synthetic datasets.}\label{tab:snr}
\end{table}

\subsection{Complex- and real-valued NN approaches}
\label{appendix:cvnnrvnn}
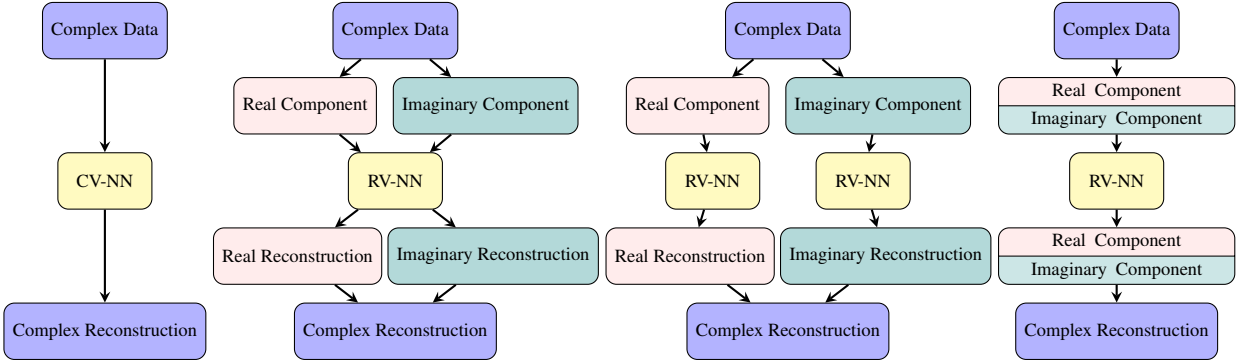
\begin{figure}[H]
    \begin{tikzpicture}[node distance=1cm]
    <TikZ code>
    \node (in1) [io] {\scriptsize Complex Data};
    \node (nn1) [rnn, below of=in1, node distance=2cm] {\scriptsize CV-NN};
    \node (out1) [io, below of=nn1, node distance=2cm] {\scriptsize Complex Reconstruction};
    \draw [arrow] (in1) -- (nn1);
    \draw [arrow] (nn1) -- (out1);
    \end{tikzpicture}
    \begin{tikzpicture}[node distance=1cm]
    <TikZ code>
    \node (in1) [io] {\scriptsize Complex Data};
    \node (blank)[blank, below of=in1] {\scriptsize };
    \node (r1) [real, left of=blank, node distance=1.2cm] {\scriptsize Real Component};
    \node (i1) [imag, right of=blank, node distance=1.2cm] {\scriptsize Imaginary Component};
    \node (nn1) [rnn, below of=blank] {\scriptsize RV-NN};
    \node (blank2)[blank, below of=nn1] {\scriptsize };
    \node (r2) [real, left of=blank2, node distance=1.3cm] {\scriptsize Real Reconstruction};
    \node (i2) [imag, right of=blank2, node distance=1.3cm] {\scriptsize Imaginary Reconstruction};
    \node (out1) [io, below of=blank2] {\scriptsize Complex Reconstruction};
    \draw [arrow] (in1) -- (r1);
    \draw [arrow] (in1) -- (i1);
    \draw [arrow] (r1) -- (nn1);
    \draw [arrow] (i1) -- (nn1);
    \draw [arrow] (nn1) -- (r2);
    \draw [arrow] (nn1) -- (i2);
    \draw [arrow] (r2) -- (out1);
    \draw [arrow] (i2) -- (out1);
    \end{tikzpicture}
    \begin{tikzpicture}[node distance=1cm]
    <TikZ code>
    \node (in1) [io] {\scriptsize Complex Data};
    \node (blank)[blank, below of=in1] {\scriptsize };
    \node (r1) [real, left of=blank, node distance=1.2cm] {\scriptsize Real Component};
    \node (i1) [imag, right of=blank, node distance=1.2cm] {\scriptsize Imaginary Component};
    \node (blank2)[blank, below of=blank] {\scriptsize };
    \node (nn1) [rnn, left of = blank2] {\scriptsize RV-NN};
    \node (nn2) [rnn, right of =blank2] {\scriptsize RV-NN};
    \node (blank3)[blank, below of=blank2] {\scriptsize };
    \node (r2) [real, left of =blank3, node distance=1.3cm] {\scriptsize Real Reconstruction};
    \node (i2) [imag, right of=blank3, node distance=1.3cm] {\scriptsize Imaginary Reconstruction};
    \node (out1) [io, below of=blank3] {\scriptsize Complex Reconstruction};
    \draw [arrow] (in1) -- (r1);
    \draw [arrow] (in1) -- (i1);
    \draw [arrow] (r1) -- (nn1);
    \draw [arrow] (i1) -- (nn2);
    \draw [arrow] (nn1) -- (r2);
    \draw [arrow] (nn2) -- (i2);
    \draw [arrow] (r2) -- (out1);
    \draw [arrow] (i2) -- (out1);
    \end{tikzpicture}
    \begin{tikzpicture}[node distance=1cm]
    <TikZ code>
    \node (in1) [io] {\scriptsize Complex Data};
    \node (conc1) [state, below of=in1] {\scriptsize Real Component \nodepart{two} \scriptsize Imaginary Component};
    \node (nn1) [rnn, below of=conc1] {\scriptsize RV-NN};
    \node (conc2) [state, below of=nn1] {\scriptsize Real Component \nodepart{two} \scriptsize Imaginary Component};
    \node (out1) [io, below of=conc2] {\scriptsize Complex Reconstruction};
    \draw [arrow] (in1) -- (conc1);
    \draw [arrow] (conc1) -- (nn1);
    \draw [arrow] (nn1) -- (conc2);
    \draw [arrow] (conc2) -- (out1);
    \end{tikzpicture}
    \caption{CV-NN and RV-NN architectures. From left to right, ComplexNet, DualReal1, DualReal2, DualReal1C. ComplexNet operates directly on complex data with complex-valued activation and loss functions.  DualReal1 splits data into real and imaginary but processes them with a single real-valued network.  DualReal2 splits into real and imaginary and processes them with two separate real-valued networks.  DualReal1C concatenates the real and imaginary components into a single real-valued vector (or two-channel vector) and then processes them with a single real-valued network which allows the model to learn unconstrained relationships between real and imaginary components.  The other models are either unable to learn these relationships at all (DualReal1 and DualReal2) or have some symmetries imposed on the relationships (ComplexNet).} \label{fig:archAE}
\end{figure}

In the main text, we focused on CV-NNs compared to one RV-NN approach working on concatenated real and imaginary components.
For the fully-connected AE, the real and imaginary components were directly concatenated into a single vector; for the convolutional CTN, the real and imaginary components were treated as two channels.
This approach allows the NN to learn relationships between the real and imaginary components of the signal.
We also investigated two other RV-NN approaches (Fig.~\ref{fig:archAE}), but found the performance was consistently worse.
In particular, we considered one RV-NN applied independently to the real and imaginary components, or separate RV-NNs applied independently to the real and imaginary components.
Neither of these approaches considers relationships between the real and imaginary components, which  could explain the poorer performance.

\subsection{Autoencoder implementation details}
\label{appendix:ae}
Four different undercomplete autoencoder architectures were implemented (Fig.~\ref{fig:archAE}). Each of these four architectures were trained in the following way.  The full 1024 dimensional space was mapped down to 10 dimensions, then 5 dimensions, before being mapped back to 10 dimensions and finally the full 1024 dimensions. Training was performed using the Adam optimizer.  Each epoch trained on the full data set.  The maximum number of training epochs was 1000 with early stopping after 5 consecutive increases in validation loss.  The loss was calculated as denoising loss, meaning the squared loss was calculated between the reproduction from the model and the clean version of the signal.  A learning rate of 0.01 was used.  The complex-valued and real-valued MSE loss function and ReLU activation function were implemented since they had the best overall performance when tested against complex ReLU, ReLU, complex cardioid, complex phase TanH, Hardtanh for activation functions and complex MSE, MSE, complex logMSE, logMSE, complex sdr, L1Loss, SNRloss for loss functions.

\subsection{ConvTasNet implementation details}
\label{appendix:ctn}
We based our implementation on the \texttt{torchaudio} implementation of ConvTasNet but with two important modifications: first, we used a deep rather than single layer encoder/decoder, and second, we developed an adaptation to use complex-valued convolutions and activation functions.
We set the depth of the encoder and decoder to three layers.
The key variation we explored in deviating from default \texttt{torchaudio} hyperparameter settings was changing the window size to 32, 64, or 128 samples.
For the complex-valued CTN, we used complex convolutions and complex PReLU activation functions to mirror the real-valued CTN~\cite{trabelsi1705deep}.
Due to GPU memory constraints, we used minibatch training with a batch size of 256 signals.
We optimized the networks for a maximum of 50 epochs with the Adam optimizer (learning rate: $3 \times 10^{-3}$), keeping the model with the highest validation error as a form of early stopping.
We found that using the log of the squared reconstruction error resulted in more stable training for CTN; this loss function is similar to an SNR-based loss.

\subsection{Baseline denoising method details}
\label{appendix:baseline}
The time-domain deconvolution method uses \texttt{least\_squares} from the Python \texttt{scipy.optimize} package to fit the generating signal model to the noisy data.
We used knowledge of the data generating distribution to set the scale and initial values of the parameters to be optimized, so that the method would have in a sense the best performance possible, but we note that in practice, the procedure can be very sensitive to these generally unknown values.
The SSA method was implemented using the Fast Gradient Cadzow approach~\cite{wang2021fast}; it works by creating a lagged time series matrix from each signal, then performing SVD, and iteratively thresholding to determine which components correspond to signal.
The wavelets baseline method processes the amplitude of the raw NQR signal using the Python package PyWavelets with sym4 wavelet, 12 decompositions, Bayes denoising, and soft thresholding~\cite{shanemd}.

\subsection{Complex- and real-valued autoencoder comparisons}
Comparison of the performance of four different autoencoders, ComplexNet, DualReal1, DualReal2, and DualReal1C across different random initialization providing insight into each model’s sensitivity to initialization.  It can be seen that DualReal1C generally has the lowest variability across seeds and the lowest MSE.  We believe it performs best due to the density of its layers, as it has the highest number of weights, and because of the flexibility in learning the relationships between the real and imaginary components of the data.  While ComplexNet is over-all outperformed by DualReal1C, more often than not, it outperforms the more na\"ive RV-NNs, DualReal1, and DualReal2.  We believe this is due to its ability to learn symmetrically constrained relationships between the real and imaginary components of the data, which DualReal1 and DualReal2 lack. Other things made apparent by \ref{tab:ae_selection} is that the sensitivity to initialization is much higher for all autoencoders across the low-SNR data sets and is much lower across the high-SNR data sets.

\begin{table}[H]
    \centering
    \setlength\tabcolsep{2.5pt}

    \begin{subtable}[t]{\columnwidth}
        \centering
        \begin{tabular}{c|c|c|c|c||c|c|c|c|}
            \multicolumn{1}{c}{} & \multicolumn{4}{c}{White noise} & \multicolumn{4}{c}{Measured noise} \\ \cline{2-9}
            & \multicolumn{2}{c|}{$f \sim 0$} & \multicolumn{2}{c||}{$f \sim F$} & \multicolumn{2}{c|}{$f \sim 0$} & \multicolumn{2}{c|}{$f \sim F$} \\ \hline
            Model & Mean & SD & Mean & SD & Mean & SD & Mean & SD \\ \hline
            ComplexNet  &  50.6 &  29.8 &  51.4 &  31.7&  48.1 &  23.6&  53.3 &  28.4\\
            DualReal1  &  61.8 &  23.5 &  77.2 &  21.3&  60.9 &  22.3&  81.4 &  22.9\\
            DualReal2   &  61.4 &  23.3 &  72.7 &  14.7&  58.0 &  22.2&  74.2 &  17.6\\
            DualReal1C  &  33.0 &  13.0 &  36.1 &  10.8&  32.1 &  12.1&  41.9 &  11.5\\
            \hline
        \end{tabular}\\
        $~~~~~~~~~~~~~~~~~~~~~~~~~~$\includegraphics[width=0.44\textwidth]{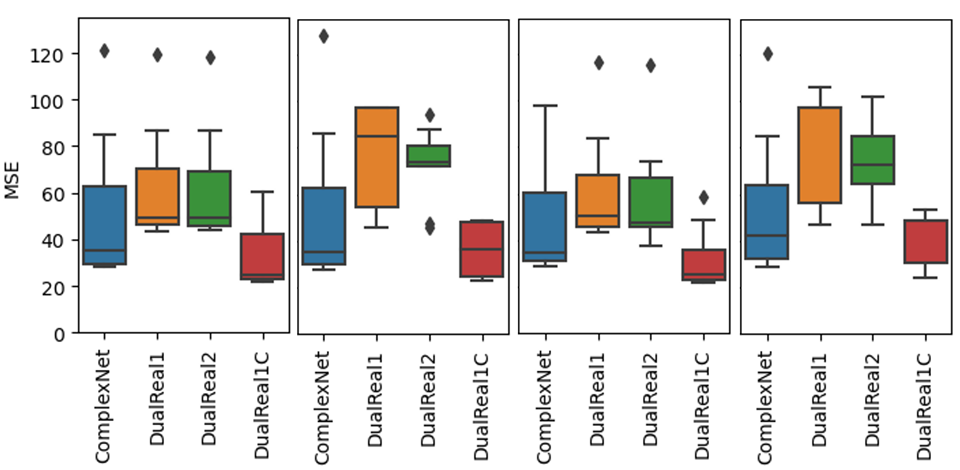}$~~~~~~~~~$
        \caption{low SNR}\label{subtab-2:ae_selection}
    \end{subtable}
    \begin{subtable}[t]{\columnwidth}
        \centering
        \begin{tabular}{c|c|c|c|c||c|c|c|c|}
            \multicolumn{1}{c}{} & \multicolumn{4}{c}{White noise} & \multicolumn{4}{c}{Measured noise} \\ \cline{2-9}
            & \multicolumn{2}{c|}{$f \sim 0$} & \multicolumn{2}{c||}{$f \sim F$} & \multicolumn{2}{c|}{$f \sim 0$} & \multicolumn{2}{c|}{$f \sim F$} \\ \hline
            Model & Mean & SD & Mean & SD & Mean & SD & Mean & SD \\ \hline
            ComplexNet  & 6.03 & 0.87 & 5.80 & 0.67& 6.51 & 0.96& 6.94 & 1.28\\
            DualReal1  & 3.16 & 0.44 & 7.95 & 0.92& 2.99 & 0.42&7.70 &1.28\\
            DualReal2   & 4.78 & 0.37& 9.96 & 0.48& 4.65 & 0.44  &  10.4 &1.34\\
            DualReal1C  & 4.96 &0.46 & 2.30 & 0.34& 5.19 & 0.43  & 2.30 & 0.23 \\
            \hline
        \end{tabular}
        $~~~~~~~~~~~~~~~~~~~~~~~~~~$\includegraphics[width=0.44\textwidth]{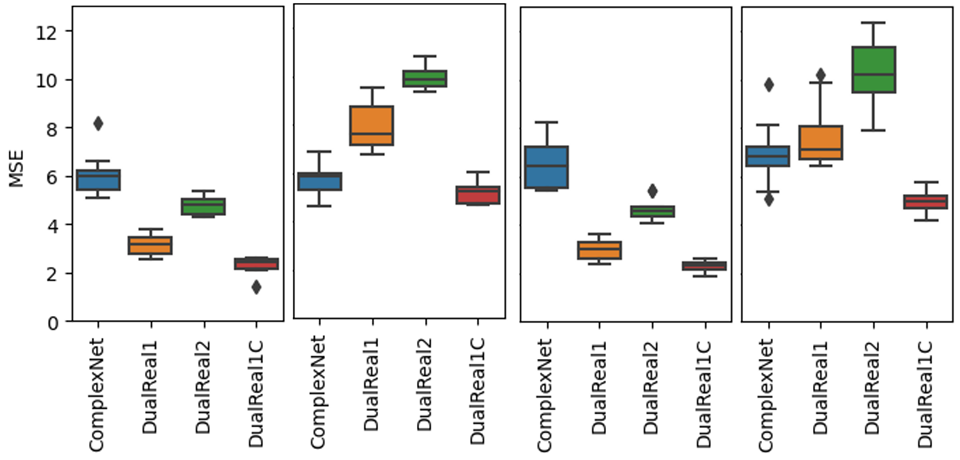}$~~~~~~~~~$
        \caption{high SNR}\label{subtab-2:ae_selection}
    \end{subtable}
    \caption{Comparing mean $\times 10^{-3}$ and standard deviation (SD $\times 10^{-3}$) of MSE from four different autoencoders.}\label{tab:ae_selection}
\end{table}

\newpage
\subsection{Complex- and real-valued ConvTasNet comparisons}
For a fixed window size (128; results similar for other window sizes), we compare the different complex- and real-valued ConvTasNet models across different seeds. 
Table~\ref{tab:ctn_cvnnrvnn} shows the mean and SD of the rescaled MSE of each model in the ensemble for the different complex- and real-valued networks across different training regimes; boxplots below the tables show the distributions of rescaled MSE across each ensemble.
In low SNR, the complex-valued CTN appears to outperform the DualReal1C CTN with smaller ensemble variation; the other real-valued approaches (DualReal1 and DualReal2) perform significantly worse.
In high SNR, the DualReal1C model appears to perform slightly better than the complex CTN.

\begin{table}[H]
    \centering
    \setlength\tabcolsep{2.5pt}

    \begin{subtable}[t]{\columnwidth}
        \centering
        \begin{tabular}{c|c|c|c|c||c|c|c|c|}
            \multicolumn{1}{c}{} & \multicolumn{4}{c}{White noise} & \multicolumn{4}{c}{Measured noise} \\ \cline{2-9}
            & \multicolumn{2}{c|}{$f \sim 0$} & \multicolumn{2}{c||}{$f \sim F$} & \multicolumn{2}{c|}{$f \sim 0$} & \multicolumn{2}{c|}{$f \sim F$} \\ \hline
            Model & Mean & SD & Mean & SD & Mean & SD & Mean & SD \\ \hline
            ComplexNet & 19.38 & 0.23 & 19.24 & 0.16 & 15.43 & 0.21 & 17.25 & 0.10 \\
            DualReal1 & 24.15 & 0.28 & 33.59 & 0.74 & 19.34 & 0.24 & 27.62 & 0.38 \\ 
            DualReal2 & 24.19 & 0.30 & 33.29 & 0.65 & 19.14 & 0.23 & 27.18 & 0.18 \\ 
            DualReal1C & 19.80 & 0.13 & 19.74 & 0.27 & 16.04 & 0.30 & 18.07 & 0.40 \\
            \hline
        \end{tabular}\\
        $~~~~~~~~~~~~~~~~~~~~~~~~~$\includegraphics[width=0.44\textwidth]{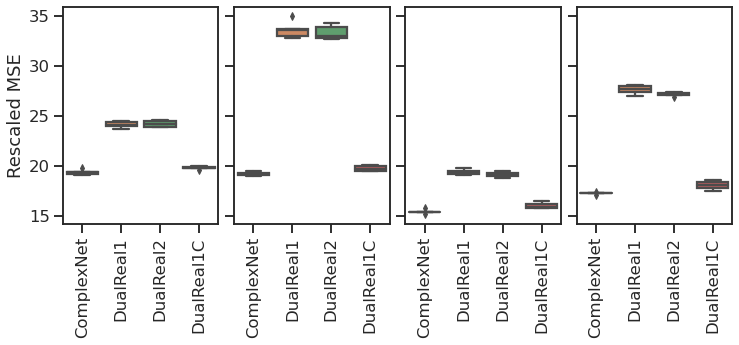}$~~~~~~~~~$
        \caption{low SNR}\label{subtab-2:ae_selection}
    \end{subtable}
    \begin{subtable}[t]{\columnwidth}
        \centering
        \begin{tabular}{c|c|c|c|c||c|c|c|c|}
            \multicolumn{1}{c}{} & \multicolumn{4}{c}{White noise} & \multicolumn{4}{c}{Measured noise} \\ \cline{2-9}
            & \multicolumn{2}{c|}{$f \sim 0$} & \multicolumn{2}{c||}{$f \sim F$} & \multicolumn{2}{c|}{$f \sim 0$} & \multicolumn{2}{c|}{$f \sim F$} \\ \hline
            Model & Mean & SD & Mean & SD & Mean & SD & Mean & SD \\ \hline
            ComplexNet & 0.67 & 0.02 & 0.66 & 0.01 & 0.56 & 0.02 & 0.62 & 0.01 \\
            DualReal1 & 0.77 & 0.02 & 1.01 & 0.01 & 0.61 & 0.02 & 0.77 & 0.01 \\ 
            DualReal2 & 0.84 & 0.02 & 1.13 & 0.01 & 0.66 & 0.01 & 0.86 & 0.01 \\ 
            DualReal1C & 0.64 & 0.01 & 0.62 & 0.01 & 0.52 & 0.01 & 0.57 & 0.02 \\ 
            \hline
        \end{tabular}
        $~~~~~~~~~~~~~~~~~~~~~~~~~$\includegraphics[width=0.44\textwidth]{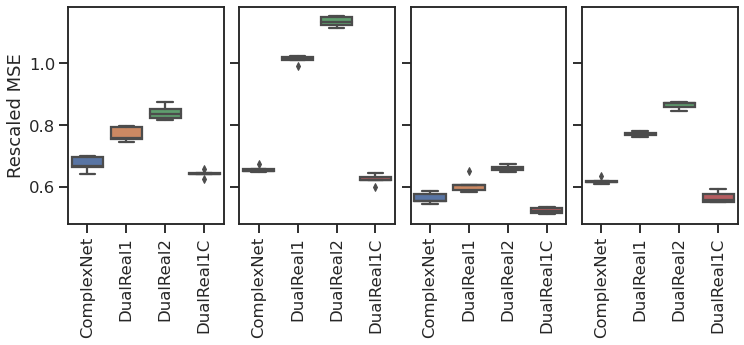}$~~~~~~~~~$
        \caption{high SNR}\label{subtab-2:ae_selection}
    \end{subtable}
    \caption{Comparing mean $\times 10^{-3}$ and standard deviation (SD $\times 10^{-3}$) of MSE from four different ConvTasNet.}\label{tab:ctn_cvnnrvnn}
\end{table}
\newpage
\subsection{ConvTasNet window selection}
Based on previous work, we determined that varying the window size in ConvTasNet would be the manipulation most likely impact model performance, so we trained models with window lengths of 32, 64, and 128 (in number of samples).
Because Table \ref{tab:nn_r2} suggests that the complex-valued ConvTasNet was better overall than the real-valued ConvTasNet across all window sizes, we here investigate how window size impacted the performance of the complex-valued ConvTasNet. 
Table \ref{subtab-1:ctn_selection} shows results for low SNR, where the table gives the mean and standard deviation of the rescaled test set MSE across the ensemble of models, with corresponding boxplots below the table showing the distribution of rescaled MSE values for each window length (across the ensemble members). In this case, it appears that the longest window length performed best across all training data sets. 
Interestingly, the opposite appeared to be true for high SNR (Table \ref{subtab-2:ctn_selection}), which suggests that smaller window sizes performed better, though the differences between window sizes were less pronounced than low SNR.
These results suggest that in high SNR, the window length is less impactful on the model performance, while in low SNR, the longer window length tends to provide better performance, perhaps because more total signal and noise information is contained in each window.

\begin{table}[H]
    \centering
    \setlength\tabcolsep{2.5pt}
    \begin{subtable}[t]{\columnwidth}
        \centering
        \begin{tabular}{c|c|c|c|c||c|c|c|c|}
            \multicolumn{1}{c}{} & \multicolumn{4}{c}{White noise} & \multicolumn{4}{c}{Measured noise} \\ \cline{2-9}
            & \multicolumn{2}{c|}{$f \sim 0$} & \multicolumn{2}{c||}{$f \sim F$} & \multicolumn{2}{c|}{$f \sim 0$} & \multicolumn{2}{c|}{$f \sim F$} \\ \hline
            Model & Mean & SD & Mean & SD & Mean & SD & Mean & SD \\ \hline
            32 & 20.60 & 0.36 & 21.15 & 0.36 & 16.50 & 0.26 & 19.02 & 0.20 \\ 
            64 & 20.74 & 0.37 & 20.79 & 0.54 & 16.87 & 0.23 & 18.67 & 0.26 \\
            128 & 19.38 & 0.23 & 19.24 & 0.16 & 15.43 & 0.21 & 17.25 & 0.10 \\
            \hline
        \end{tabular}
        $~~~~~~~~~~~~~~~~$\includegraphics[width=0.44\textwidth]{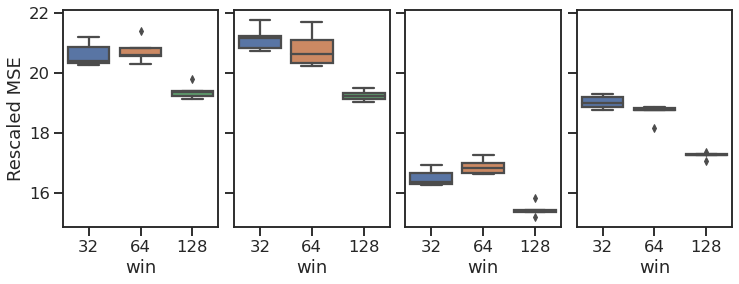}$~~~~~~~~~$
        \caption{low SNR}\label{subtab-1:ctn_selection}
    \end{subtable}
    \begin{subtable}[t]{\columnwidth}
        \centering
        \begin{tabular}{c|c|c|c|c||c|c|c|c|}
            \multicolumn{1}{c}{} & \multicolumn{4}{c}{White noise} & \multicolumn{4}{c}{Measured noise} \\ \cline{2-9}
            & \multicolumn{2}{c|}{$f \sim 0$} & \multicolumn{2}{c||}{$f \sim F$} & \multicolumn{2}{c|}{$f \sim 0$} & \multicolumn{2}{c|}{$f \sim F$} \\ \hline
            Model & Mean & SD & Mean & SD & Mean & SD & Mean & SD \\ \hline
            32 & 0.67 & 0.04 & 0.61 & 0.02 & 0.52 & 0.03 & 0.55 & 0.02 \\ 
            64 & 0.68 & 0.04 & 0.65 & 0.03 & 0.55 & 0.01 & 0.56 & 0.01 \\ 
            128 & 0.67 & 0.02 & 0.66 & 0.01 & 0.56 & 0.02 & 0.62 & 0.01 \\ 
            \hline
        \end{tabular}
        $~~~~~~~~~~~~~$\includegraphics[width=0.45\textwidth]{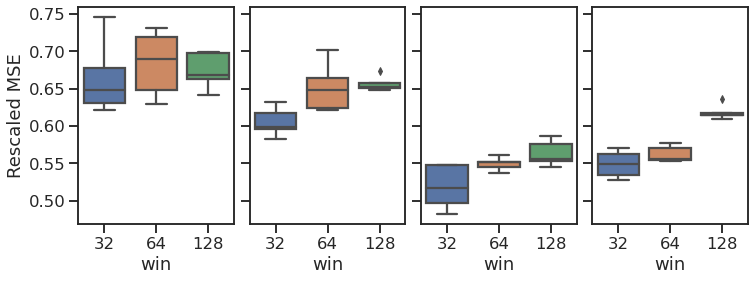}$~~~~~~~~~$
        \caption{high SNR}\label{subtab-2:ctn_selection}
    \end{subtable}
    \caption{Comparing mean $\times 10^{-3}$ and standard deviation (SD $\times 10^{-3}$) of MSE from different complex ConvTasNet window sizes.}\label{tab:ctn_selection}
\end{table}

\subsection{Out-of-distribution data and results}
\label{appendix:ood}
We generated out-of-distribution (OOD) test sets in two ways.
First, we applied trained neural networks (NNs) to test sets that had different noise distributions (trained on white noise, applied to data with measured noise, or vice versa) with all other data characteristics held fixed.
To measure performance, we calculated a standardized mean squared error (MSE), in which each true signal and prediction was first rescaled by the true signal amplitude $A$.
As a measure of ensemble uncertainty, we calculated the standard deviation (SD) in the individual model MSE values.
Table \ref{tab:ood1} shows that the NN performance in terms of ensemble mean squared error (MSE) remains stable when the noise distribution changes, and that the ensemble standard deviation (SD) also remains stable.
These results suggest that the networks are robust to some changes in the noise distribution at deployment.

\begin{table}[ht]
    \begin{subtable}[t]{\columnwidth}
        \centering
        \begin{tabular}{|l|l|c|c|c|c|}\hline
            SNR & $f$ &      w $\xrightarrow{}$ w &      w $\xrightarrow{}$ m &      m $\xrightarrow{}$ m &      m $\xrightarrow{}$ w \\ \hline
            high & 0 &  0.54(0.04) &  0.44(0.03) &  0.44(0.03) &  0.56(0.03) \\
            high & $F$ &  0.50(0.02) &  0.48(0.02) &  0.45(0.02) &  0.51(0.02) \\
            low & 0 & 17.9(0.23) & 14.6(0.19) & 14.1(0.21) & 18.3(0.28) \\
            low & $F$ & 17.8(0.16) & 17.1(0.26) & 15.8(0.10) & 18.2(0.17) \\ \hline
        \end{tabular}
        \caption{Ensemble MSE(SD) $\times 10^{-3}$  for CTN (CV) models.}
        \label{subtab-1:ood1_ctn}
    \end{subtable}
    \begin{subtable}[t]{\columnwidth}
        \centering
        \begin{tabular}{|l|l|c|c|c|c|}\hline
            SNR & $f$ &      w $\xrightarrow{}$ w &      w $\xrightarrow{}$ m &      m $\xrightarrow{}$ m &      m $\xrightarrow{}$ w \\ \hline
            high & 0 & 1.86(0.25) & 1.81(0.16) & 1.89(0.13) & 1.95(0.12)\\
            high & $F$ & 4.36(0.22) & 4.28(0.23) & 4.10(0.35) & 4.17(0.34)\\
            low & 0 & 22.5(5.68) & 22.3(0.06) & 21.8(0.87) & 22.4(8.44)\\
            low & $F$ & 23.6(4.83) & 23.4(5.28) & 29.3(6.34) & 29.5(5.75)\\ \hline
        \end{tabular}
        \caption{Ensemble MSE(SD) $\times 10^{-3}$  for AE (DualReal1C) models.}
        \label{subtab-1:ood1_ae}
    \end{subtable}
    \caption{Ensemble MSE(SD) $\times 10^{-3}$ for NN models trained under four SNR/central frequency conditions (rows); columns indicate training and test noise distributions, where `w' is white noise and `m' is measured noise. Here, w $\rightarrow$ w indicates that the model was trained with white noise and applied to test data with white noise, while w $\rightarrow$ m indicates the model was trained with white noise and applied to test data with measured noise (OOD), and similar for the remaining two columns. Generally, moving from in-distribution to OOD test sets does not increase MSE or ensemble uncertainty, indicating robustness to noise distribution.} \label{tab:ood1}
\end{table}

Second, we generated test sets that had different signal frequency distributions ($f \sim 300$Hz or $f \sim 1900$Hz) with the frequency of an individual example drawn as Uniform($f-150, f+150$).
This means that during test time the signal frequencies are shifted so that they do not overlap the frequencies seen during training, but are still relatively close to the training frequencies.
Table \ref{tab:ood2} shows that performance degrades substantially in this case.
In high SNR, the CTN ensemble SD increases, suggesting that the ensemble uncertainty could help indicate OOD data, but this pattern does not hold in low SNR.
This result suggests that training sets should be designed to contain the range of frequencies anticipated during deployment.

\begin{table}[ht]
    \centering
    \setlength\tabcolsep{2.5pt}
    \begin{subtable}[t]{\columnwidth}
        \centering
        \begin{tabular}{c|c|c|c|c||c|c|c|c|}
            \multicolumn{1}{c}{} & \multicolumn{4}{c}{White noise} & \multicolumn{4}{c}{Measured noise} \\ \cline{2-9}
            Trained & \multicolumn{2}{c|}{$f \sim 0$} & \multicolumn{2}{c||}{$f \sim 1600$} & \multicolumn{2}{c|}{$f \sim 0$} & \multicolumn{2}{c|}{$f \sim 1600$} \\ \cline{2-9}
            Tested & $f \sim 0$ & $f \sim 300$ & $f \sim 1600$ & $f \sim 1900$  & $f \sim 0$ & $f \sim 300$  & $f \sim 1600$ & $f \sim 1900$  \\ \hline
            AE & 1.86(0.25) & 29.5(0.32) &  4.36(0.22) & 35.0(0.18) & 1.89(0.13) & 29.7(0.09) & 4.10(0.35) & 34.7(0.15) \\
            CTN & 0.54(0.04) & 16.3(1.26) &  0.50(0.02) & 21.5(3.44) & 0.44(0.03) & 14.7(1.42) &  0.45(0.02) & 18.1(4.09) \\
            \hline
        \end{tabular}
        \caption{Relative MSE for out of distribution high-SNR data.}\label{subtab-1:ood2}
    \end{subtable}
    \begin{subtable}[t]{\columnwidth}
    \centering
        \begin{tabular}{c|c|c|c|c||c|c|c|c|}
            \multicolumn{1}{c}{} & \multicolumn{4}{c}{White noise} & \multicolumn{4}{c}{Measured noise} \\ \cline{2-9}
            Trained & \multicolumn{2}{c|}{$f \sim 0$} & \multicolumn{2}{c||}{$f \sim 1600$} & \multicolumn{2}{c|}{$f \sim 0$} & \multicolumn{2}{c|}{$f \sim 1600$} \\ \cline{2-9}
            Tested & $f \sim 0$ & $f \sim 300$ & $f \sim 1600$ & $f \sim 1900$  & $f \sim 0$ & $f \sim 300$  & $f \sim 1600$ & $f \sim 1900$  \\ \hline
            AE & 22.5(5.68) & 47.1(1.41) & 23.6(4.83)  & 48.6(1.13) & 21.8(0.87) & 47.1(1.32) & 29.3(6.34) & 48.4(1.16) \\
            CTN & 17.9(0.23) & 65.2(0.24) & 17.8(0.16) & 65.9(0.45) & 14.1(0.21) & 59.5(0.26) & 15.8(0.10) & 63.2(0.67)  \\
            \hline
        \end{tabular}
        \caption{Relative MSE for out of distribution low-SNR data.}\label{subtab-2:ood2}
    \end{subtable}
    \caption{Ensemble MSE(SD) $\times 10^{-3}$ for NN models trained with central frequencies near 0 and 1600Hz applied to test data at 0 or 1600Hz (in-distribution) or 300 or 1900Hz (OOD). Moving from in-distribution to OOD signals results in higher MSE across both NN models. In high SNR, the ensemble uncertainty increases when moving to OOD data, suggesting it could be used as an indicator of OOD data during deployment. However, the same pattern does not hold for low SNR data.}\label{tab:ood2}
\end{table}

\end{document}